\documentclass[sigconf]{acmart}

\usepackage{graphicx} 
\usepackage{subcaption}
\usepackage{algorithm}
\usepackage{algpseudocode}
\usepackage{algorithmicx}
\usepackage{multirow}
\usepackage{tikz}
\usepackage{color, soul}
\soulregister\cite7
\soulregister\ref7
\soulregister\pageref7
\usepackage{bm}
\usepackage{threeparttable}

\renewcommand\footnotetextcopyrightpermission[1]{} 

\graphicspath{{images/}} 
  
\begin{document}
\title{KD\texorpdfstring{$k$}: A Defense Mechanism Against Label Inference Attacks in Vertical Federated Learning}

\author{Marco Arazzi}
\affiliation{%
	\institution{University of Pavia}
	\streetaddress{}
	\city{Pavia}
	\state{}
	\country{Italy}
	\postcode{}
}
\email{marco.arazzi01@universitadipavia.it}

\author{Serena Nicolazzo}
\affiliation{%
	\institution{University of Milan}
	\streetaddress{}
	\city{Milan}
	\country{Italy}}
\email{serena.nicolazzo@unimi.it}

\author{Antonino Nocera}

\affiliation{%
	\institution{University of Pavia}
	\streetaddress{}
	\city{Pavia}
	\state{}
	\country{Italy}
	\postcode{}
}
\email{antonino.nocera@unipv.it}

\begin{abstract}
Vertical Federated Learning (VFL) is a category of Federated Learning in which models are trained collaboratively among parties with vertically partitioned data.
Typically, in a VFL scenario, the labels of the samples are kept private from all the parties except for the aggregating server, that is the label owner. Nevertheless, recent works discovered that by exploiting gradient information returned by the server to bottom models, with the knowledge of only a small set of auxiliary labels on a very limited subset of training data points,
an adversary can infer the private labels. These attacks are known as label inference attacks in VFL. In our work, we propose a novel framework called KD$k$, that combines Knowledge Distillation and $k$-anonymity to provide a defense mechanism against potential label inference attacks in a VFL scenario. 
Through an exhaustive experimental campaign we demonstrate that by applying our approach, the performance of the analyzed label inference attacks decreases consistently, even by more than $60\%$, maintaining the accuracy of the whole VFL almost unaltered.
\end{abstract}

\keywords{Federated Learning, Vertical Federated Learning,  VFL, Label Inference Attack, Knowledge Distillation, \texorpdfstring{$k$}-anonymity.}
  
\maketitle
\thispagestyle{empty}
\pagestyle{plain}

\section{Introduction}
\balance
Federated Learning (FL, for short) has emerged in the last years as a key technology enabling collaborative model training across different entities without the need to gather data in a central location \cite{mcmahan2017communication}. The application of this paradigm is advantageous when organizations or individuals have to cooperate on model development without revealing their sensitive data.
Since model updates are performed locally also communication costs are reduced, moreover, heterogeneous data can be integrated more easily maintaining the peculiarities of the different participants and building joint and richer data pools.
Nevertheless, although this approach is designed to keep raw data on local devices, there is still a risk of indirect information leakage, particularly during the model aggregation stage.

According to the different data partition strategies adopted, three main categories of FL have been formulated, i.e., Horizontal, Vertical, and Transfer Learning \cite{yang2019federated}.
Horizontal FL (HFL, for short) requires all parties to hold the same attribute space but different sample space and it is suitable for scenarios where various regional branches of the same business want to build a richer dataset. Vertical FL (VFL, hereafter), instead, is based on the collaborations
among non-competing entities with vertically partitioned data that share overlapping data samples but differ in the feature space (i.e., a mobile phone company and a TV streaming service provider). Finally, Transfer Learning is applicable for scenarios in which there is little overlapping in data samples and features, and multiple subjects with heterogeneous distributions build models collaboratively.
Hence, HFL is characterized by independent
training and no raw data is shared among participants, only model updates, reducing the risk of privacy concerns and the overhead of transmitted data.
VFL, instead, exploits deeper attribute dimensions and leads to more accurate models because data features are complementary across different sources. This results in a higher complexity, also due to feature-level coordination. 

However, even if raw data is not shared due
to the calculation and exchange of features, combining information across features and the possible presence of a compromised participant may raise privacy leakage \cite{wei2022vertical,vucinich2023current}.
Possible attacks that represent a significant concern in this context are the {\em Label Inference Attacks}, because of the high sensitivity of the labels that may reveal crucial clients' information (e.g., diseases, financial details, etc.).
In our work, we start considering the label inference attacks to VFL described in the recent paper of Fu et al. \cite{fu2022label}. 

In our study, we provide a defense against the above-cited attacks.
Our application scenario is the classical VFL scenario in which two typologies of participants, a server and a set of clients, collaboratively train an ML holding different feature spaces. The server or active participant also stores the labels that are kept private from the other passive participants. The adversary controls some passive participants and aims at discovering the private labels.
The authors of \cite{fu2022label} demonstrate that, in this scenario, different types of label inference attacks succeed in most cases reaching good accuracy results. Specifically, in \cite{fu2022label}, Fu et al. discover that thanks to {\em (i)} the trained local model held by the malicious participant, and {\em (ii)} the received gradients of the loss that contains hidden information about labels, the attacker can succeed in conducting a label inference attack.
They describe three possible categories of label inference attacks. The first attack is a {\em passive attack}, in which, with the help of some auxiliary labeled data, the malicious participant can fine-tune his/her trained bottom model to infer the labels in a semi-supervised manner. 
The second type is an {\em active attack}, in which the attacker tries to scale up the learning rate of her/his bottom model during the training phase to force the top model to rely more on her/his model thus boosting the label inference accuracy. The third type is a {\em direct attack}
through which the adversary can infer labels by analyzing the signs of gradients from the server. 

Our proposal consists of designing a novel framework called KD$k$ (Knowledge Discovery and $k$-anonymity) as a defense mechanism relying on an additional component for the server (or active) participant. This includes a {\em Knowledge Distillation} step and an {\em obfuscation} algorithm. Specifically, 
Knowledge Distillation (KD, hereafter) \cite{gou2021knowledge} is an ML
compression technique able to transfer knowledge from a larger teacher model to a smaller student one. The teacher network produces softer probability distributions instead of hard labels that can better capture essential features and relationships in the data.

We include in the active participant a teacher network whose outputs are soft labels. These are then processed by an algorithm based on the concept of $k$-anonymity \cite{samarati1998protecting} to add a further level of uncertainty. This step groups together the $k$ labels with the higher probabilities making it hard for the attacker to infer the most probable one. Then the top model of the server can be fed with these new soft and partly anonymized labels and the VFL tasks can be executed collaboratively.

Our experimental campaign demonstrates that using our approach the accuracy of the three types of label inference attacks decreases significantly.

In summary, the main contributions of this paper are:
\begin{itemize}
    \item we design a countermeasure for the different types of label inference attacks proposed by \cite{fu2022label}.

    \item We conduct an experimental campaign to demonstrate that the accuracy of all the analyzed types of label inference attacks consistently decreases if our complete approach is applied.
    
    \item We provide a comparison with existing defense strategies and show the higher effectiveness of our solution.
\end{itemize}

The organization of this paper is outlined as follows. Section \ref{sec:related_work} describes the main works related to our approach. Section \ref{sec:background} delves
into the details about FL, $k$-anonimity, and Knowledge Distillation that are essential to the understanding of our solution. Section \ref{sec:labelAttack} presents the types of label inference attacks against which we provide a defense. Section \ref{sec:experimental} discusses the experimental
campaign, including the setup and results of our defense mechanisms. Ultimately, Section \ref{sec:conclusion} concludes the work and presents possible future directions. 

\section{Related Work}
\label{sec:related_work}

Recent works have shown that FL is vulnerable to multiple types
of inference attacks, such as membership inference,
property inference, and feature inference \cite{melis2019exploiting,zhu2019deep,nasr2018comprehensive}.
The objective of a membership inference is to discriminate whether a specific
record is in a party's training dataset or not. Nevertheless, this type of attack has no reason to exist in
VFL as every participant knows all the training sample IDs.
Property inference aims to extract some properties about a party's training dataset,
which are uncorrelated to the training task. In a feature
inference attack, instead, a party tries to recover the samples used in another party's
training dataset. For instance, Luo et al. propose a feature
inference attack for VFL \cite{luo2021feature}, in which the active party tries to infer the features owned by the passive party. However, the authors
strongly assume that the active party knows the model parameters of the passive party, which is difficult to achieve in
real-world scenarios. 
Differently from the above-cited works, our proposal deals with a different type of inference attack in VFL, known as a label inference attack, conducted by the passive party and aiming at leaking the labels owned by the active participant. Since labels often contain highly sensitive information this type of attack deserves more and more attention. Although finding possible defense strategies against these attacks is crucial they are still an open challenge and only a few efforts have been made.

For example, the articles \cite{li2022label,liu2022clustering} study label inference attacks in VFL, but they specifically focus on split learning scenarios. In particular, \cite{li2022label} formalizes a threat model for label leakage in two-party split learning in the context of binary classification and proposes a countermeasure based on random perturbation techniques that minimize the
amount of label leakage of a worst-case adversary. Whereas, the proposal in \cite{liu2022clustering} presents a passive clustering label inference attack for split learning, in which the adversary
(that can be any clients or the server) retrieves the private
labels by collecting the exchanged gradients and smashed
data both
during and after the training phase. \cite{liu2022batch} design the inversion and replacement attacks to disclose private labels from batch-level messages in a VFL whose communication is protected by a Homomorphic encryption mechanism and a confusional autoencoder (CoAE) method as a possible countermeasure.
The proposal in \cite{sun2022label} deals with the design of a label leakage attack from the forward
embedding in two-party split learning and a corresponding defense that reduces the
distance correlation between cut layer embedding and private labels.
Kholod et al., \cite{kholod2021parallelization} propose a parallelization method to decrease data transmission, and, consequently, both the
learning cost and privacy leakage risk.
A framework called LabelGuard has been designed in \cite{xia2023cascade} to defend against label inference attacks via a cascade VFL algorithm through a minimization of the VFL task training loss.

In this work, we start from the proposals \cite{fu2022label}. The authors of \cite{fu2022label}
describe three kinds of label inference attack, i.e., passive label inference attack, active label inference attack, and direct
label inference attack, for VFL. Adversaries could infer the labels of the active party
from both the received plaintext gradients and obtained plaintext final model weights. Although these attacks
are very effective, they make a strong assumption on auxiliary labels that have to be held for the adversary.

\section{Background}
\label{sec:background}
In this section, we describe some concepts useful to understand our approach. In particular, we examine the key aspects and different categories of Federated Learning, we recall the concept of $k$-anonimity and we delve into the analysis of the main features of Knowledge Distillation.

Table \ref{tab:SystemSymbols} summarizes the acronyms used in this paper.

\begin{table}
\centering
  \caption{Summary of the acronyms used in the paper.\label{tab:SystemSymbols}}
  \begin{tabular}{ll}
\hline
    Symbol&Description\\
\hline
    DL & Deep Learning\\
    FCNN & Fully Connected Neural Network\\
    FL & Federated Learning\\
    FTL & Federated Transfer Learning\\
    GC & Gradient Compression\\
    HFL & Horizontal Federated Learning\\
    KD & Knowledge Distillation\\
    ML & Machine Learning\\
    NG & Noise Gradient\\
    OA & Original Architecture \\
    PPDL & Privacy-Preserving Deep Learning \\
    VFL & Vertical Federated Learning\\
\hline
\end{tabular}
\end{table}

\subsection{Federated Learning}
\label{subsec:vfl}

FL is a Machine Learning method designed to train a model in a distributed manner across different devices holding local data samples. The fact that data is not transferred and centralized is advantageous for privacy preservation reasons and network traffic reduction due to large data volumes. 

The actors of this protocol are $\mathcal{C}$ devices (or ``clients'' or ``workers''), running local training and holding private data; and a central server called ``aggregator'', that coordinates the whole FL process aggregating the local updates. 
Specifically, FL aims to train a global model $\mathbf{w}$ by uploading the weights of local models 
$\{\mathbf{w}^i|i \in \mathcal{C}\}$ to a parametric server optimizing a loss function:

\begin{equation}
    \min\limits_{\mathbf{w}} l(\mathbf{w}) = \sum_{i=1}^n{\frac{s_i}{\mathcal{C}}L_i(\mathbf{w}^i)}
\end{equation}
\noindent
where $L_i(\mathbf{w^i})= \frac{1}{s_i}\sum_{j \in I_i}{l_j(\mathbf{w}^i, x_i)}$ is the loss function, $s_i$ is the local data size of the {\em i}-th worker, and $I_i$ identifies the set of data indices
with $|I_i|=s_i$, and $x_j$ is a data point.

\begin{figure}
    \centering
    \includegraphics[width=0.48\textwidth]{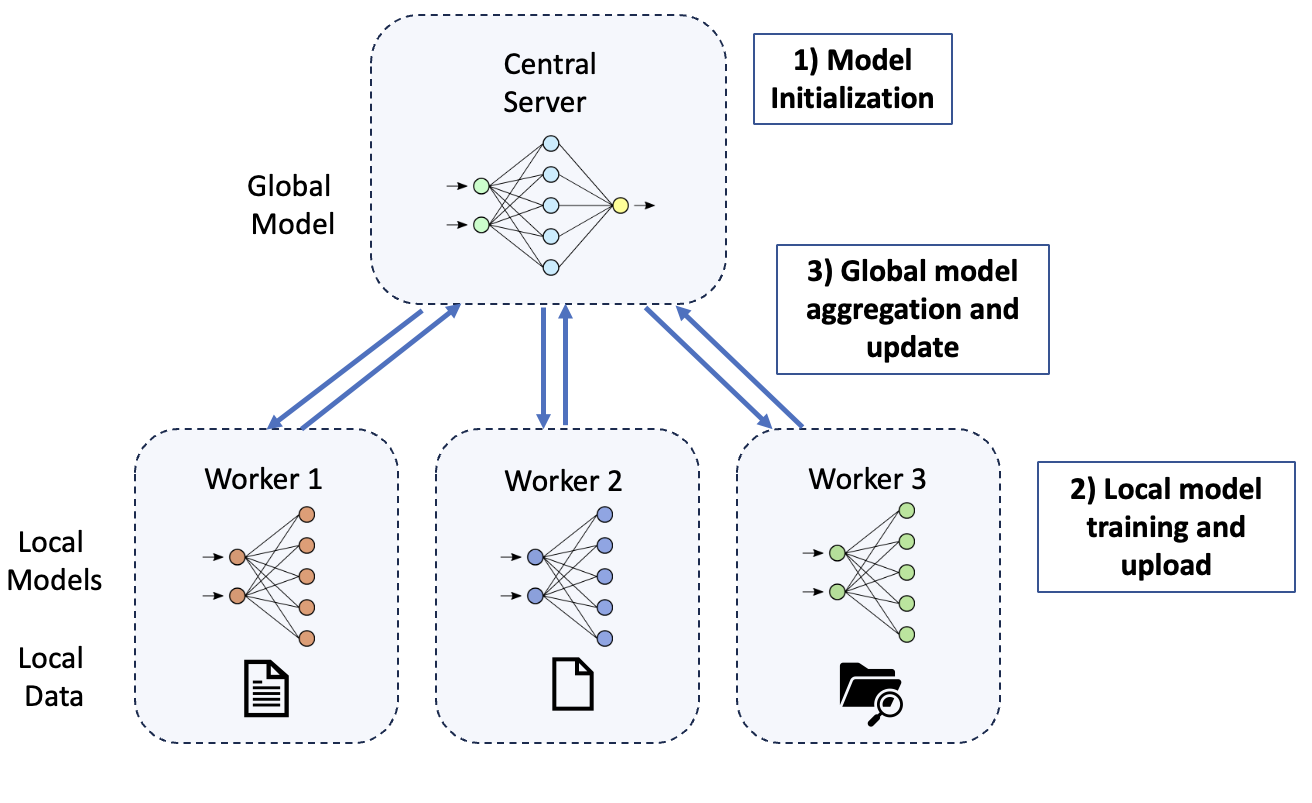}
    \caption{The Federated Learning workflow \label{fig:workflowFL}}
    \Description[]{}
\end{figure}

The basic FL workflow, shown in Figure \ref{fig:workflowFL}, can be divided into the following steps \cite{zhang2021survey}:

\begin{enumerate}
    \item Model initialization, in which the central server initializes all the necessary parameters for the global ML model $\mathbf{w}$. This phase also includes the workers' selection process. 
    \item Local model training and upload, in which the workers download the current global model and perform local training using their private data. After that, each client computes the model parameter updates and transmits them to the central server. The local training typically involves multiple iterations of gradient descent, back-propagation, or other optimization methods to improve the local model's performance. Specifically, at the {\em t}-iteration, each client updates the global model by training with their datasets: $\mathbf{w}^i_t \leftarrow \mathbf{w}^i_t - \eta \frac{\partial L(\mathbf{w}_t,b)}{\partial \mathbf{w}^i_t}$, where $\eta$ and $b$ identify the learning rate and local batch, respectively.
    \item Global model aggregation and update, in which the central server collects and aggregates the model parameter updates from all the workers, $\{\mathbf{w}^i|i \in \mathcal{C}\}$. The central server can employ various aggregation methods like averaging, weighted averaging, or secure multi-party computation to incorporate the received updates from each client, thus improving the performance of the global model.
\end{enumerate}

\begin{figure}
\centering
\begin{subfigure}{0.33\textwidth}
    \includegraphics[width=\textwidth]{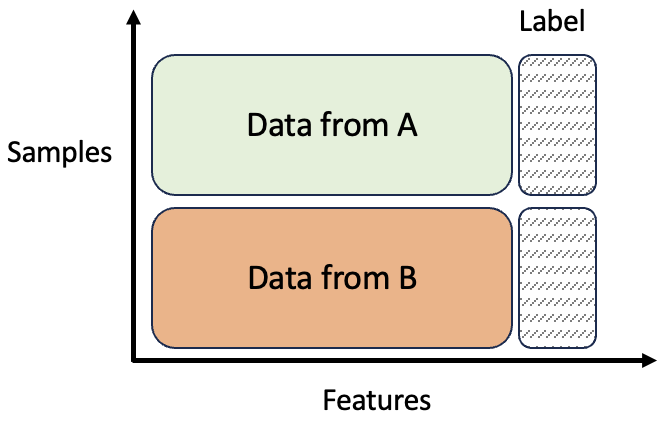}
    \caption{Horizontal FL (HFL) }
    \label{fig:hfl}
\end{subfigure}
\hfill
\begin{subfigure}{0.33\textwidth}
    \includegraphics[width=\textwidth]{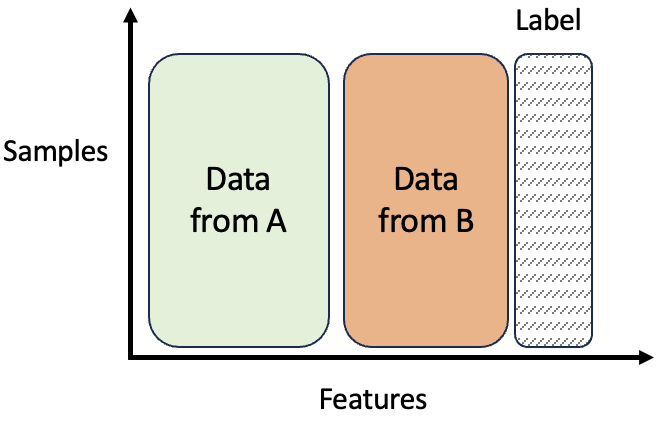}
    \caption{Vertical FL (VFL)}
    \label{fig:vfl}
\end{subfigure}
\hfill
\begin{subfigure}{0.33\textwidth}
    \includegraphics[width=\textwidth]{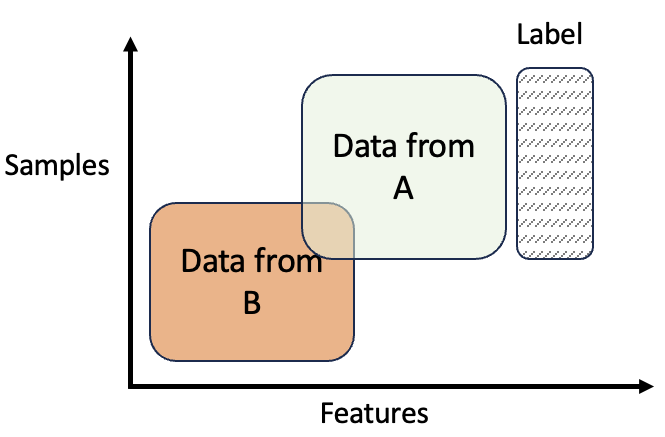}
    \caption{Federated Transfer Learning (FTL)}
    \label{fig:ftl}
\end{subfigure}
\caption{The three categories of FL divided for feature and sample spaces}\label{fig:HorizontaleVertivalFTL}
\Description[]{}
\end{figure}

FL can be classified into three scenarios according to the different data partition strategies adopted, i.e., Horizontal, Vertical, and Transfer Learning types \cite{yang2019federated}, as shown in Figure \ref{fig:HorizontaleVertivalFTL}.

As visible in Figure \ref{fig:hfl}, \textbf{Horizontal FL} (HFL, hereafter), or sample-based FL, is introduced in the scenarios in which the dataset of the parties share the same feature space, but have different spaces in samples. For instance, two branches of the same company (two regional banks or two hospitals) hold data referring to users of distinct areas, but they share the same feature spaces (i.e., the same characteristics). If the two parties aggregate their samples, they could build a larger dataset and then train a more accurate model. Yet, privacy law forbids the direct sharing of sensitive user data. In such cases, HFL can help to solve this challenge and provide a rich privacy-preserving dataset.

\textbf{Vertical FL} (VFL, for short) or feature-based FL, instead, applies to the case where the datasets share overlapping data samples but differ in the feature space (as shown in \ref{fig:vfl}. Consider the case of two different businesses sharing information about overlapping sets of customers. For example, a mobile phone company and a TV streaming service provider can have common clients, but the types of data related to them are very different. 
Nevertheless, a rich prediction model for service purchases can be built leveraging both datasets collaboratively, but directly revealing personal customer information to third parties is not always possible because of GDPR.  Hence, VFL can represent a solution for this setting allowing the company to collaboratively train a join model with a rich dataset.
VFL can be with or without model splitting:
\begin{itemize}
    \item In the presence of model splitting, every participant runs a bottom (or local) model without sharing the entire model with other participants and relying on features locally available at each party. The final top (or global) model is reconstructed by a server that combines the locally trained model portions to compute a final output. For instance, one party may focus on training the model's parameters related to demographic features, while another party may work on parameters related to data about purchase.
    \item In the absence of model splitting, the model remains centralized, and each party calculates gradients of the loss relying on its local data, then it shares these gradients with a central server, which aggregates them to update the global model.
\end{itemize}

After the training process is finished, at the inference time, VFL requires all participants
to get involved, instead for HFL, the trained global model is shared with every participant which performs inference individually.

The last category of FL is \textbf{Federated Transfer Learning} (FTL, hereafter) which is a combination of the two previous types \cite{weiss2016survey}. Indeed, FTL is applicable for scenarios in which there is little overlapping in both data samples and features as visible in Figure \ref{fig:ftl}. For instance, the case in which multiple subjects with heterogeneous distributions build models collaboratively. Consider the case where an American company producing health IoT sensors wants to join its data sample with a private health clinic in Canada.
These two entities have to follow law restrictions. Moreover, both their sets of clients and features have small intersections. In this case, FTL techniques can be applied
to provide a solution and make the two parties cooperate in the building of one model.

\subsection{k-Anonimity}
\label{subsec:kanon}

The concept of $k$-anonymity, first described in \cite{samarati1998protecting}, represents one foundational principle in database theory for privacy-preserving data publishing.
It aims to safeguard the anonymity of the individuals' data by ensuring that each record in the dataset is indistinguishable from at least $k-1$ other records. 
Several procedures can be applied to attributes to obtain $k$-anonymity, such as:
\begin{itemize}
    \item Suppression, which implies removing or cleansing certain information.
    \item Generalization is replacing distinctive values with more general ones (e.g., substituting exact ages with age ranges).
\end{itemize}

\subsection{Knowledge Distillation}
\label{subsec:kdistillation}

Knowledge Distillation (KD, for short) is an ML model compression technique, in which the knowledge from a complex model, or ``teacher'' model, is transferred to a smaller and more efficient model, known as the ``student'' model without a significant drop in accuracy \cite{gou2021knowledge}.
The general idea was first presented by Bucilua et al. in 2006 \cite{bucilua2006model} and modeled in its current known form in 2014 by Hinton et al. \cite{hinton2015distilling} who found it easier to train a classifier using the outputs of another classifier as target values than using actual ground-truth labels. The teacher network outputs are represented by the so-called soft probabilities that contain more information about a data point than just the class label (or hard predictions) and are the input of the student network. 

In practice, given an input $x$ the teacher network produces a vector of scores $s_x^t=[s_1^t,s_2^t,\dots,s_K^t]$ that are converted into probabilities:
\begin{equation}
    p_k^t(x)=\frac{e^{s_k^t}}{\sum_je^{s_j^t}}
\end{equation}

Hinton et al. \cite{hinton2015distilling} proposed to modify these probabilities in {\em soft probabilities} as following:

\begin{equation}
    p_k^t(x)=\frac{e^{s_k^t}/\tau}{\sum_je^{s_j^t}/\tau}
\end{equation}

\noindent
where $\tau$ is a hyperparameter.
A student network will produce a softened class probability
distribution, $\tilde{\mathbf{p}}^s(x)$. The loss for the student network is a linear combination of the cross entropy loss, namely $\mathcal{L}_{cl}$ and  a knowledge distillation loss $\mathcal{L}_{KD}$:

\begin{equation}
   \mathcal{L} = \alpha\mathcal{L}_{cl} - (1 - \alpha) \mathcal{L}_{KD}
\end{equation}

\noindent
where $\mathcal{L}_{KD}= -\tau^2\sum_k{\tilde{p}^t(x)\log{\tilde{p}^s(x)}}$ and $\alpha$ and $\tau$ are hyperparameters.

\begin{figure}
    \centering
    \includegraphics[width=0.5\textwidth]{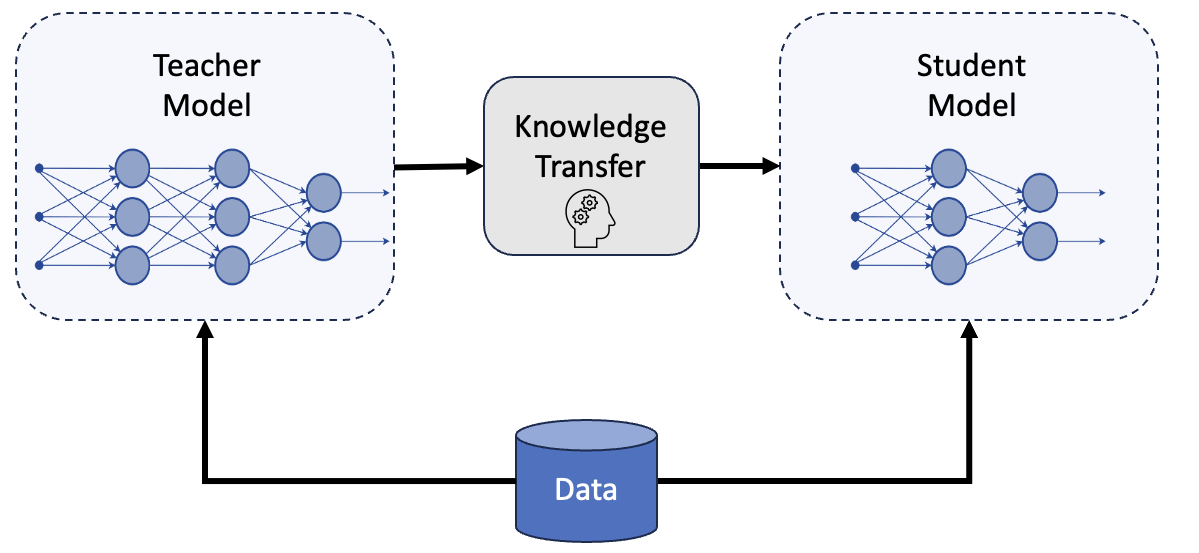}
    \caption{Generic architecture of knowledge distillation using a teacher-student model}\label{fig:kdistillation}
    \Description[]{}
\end{figure}

Figure \ref{fig:kdistillation} shows the generic architecture of the KD using the teacher-student model. Thanks to the distillation algorithm the student mimics the teacher network learning the relationship between different classes discovered by the teacher model that contains information beyond the ground truth labels. 

\section{Label Inference Attacks}
\label{sec:labelAttack}
In this section, we describe the most common types of label inference attacks against VFL, which we focus on to design our defense strategy.
As typically done in the related literature, we make explicit reference to the more complex scenario in which VFL is combined with model splitting \cite{fan2023flsg} (see Section \ref{sec:background}).

In this setting, as originally proposed by \cite{fu2022label}, label inference attacks are carried out by adversaries, controlling one or more of the bottom models, which aim to infer the private labels for any samples in the dataset. Recall that, according to the model splitting paradigm, only the active party, i.e., the server, has the classification layer, whose objective is the prediction of the correct label for each datapoint in input. Therefore, labels are available only to this active party of the FL system and, therefore, are considered sensitive information.
To carry out a label inference attack, adversaries can mainly exploit two main aspects of VFL that, according to \cite{fu2022label}, may generate label leakage, namely: 

\begin{itemize}
    \item the trained local model that is under the full control of the malicious participant;
    \item the received gradients of the loss that contain hidden information about labels.
\end{itemize}

In the following, we describe four main types of label inference attacks, namely: {\em (i)} Passive Label Inference attack \cite{fu2022label}; {\em (ii)} Active Label Inference attack \cite{fu2022label}; and {\em (iii)} Direct Label Inference Attack \cite{fu2022label}. 

\subsection{Passive Label Inference Attack}
\label{sub:passiveAttack}

Adversaries can perform this attack by exploiting their locally owned bottom model.
It is referred to as {\em passive} because the malicious participant does not perform any active action during the training or inference phase, but she/he remains {\em honest but curious}.
This type of attack assumes that the adversarial can rely on a few auxiliary labeled data (in \cite{fu2022label} only the $0.08\%$ of the labeled training samples have been used as auxiliary labels in the experimental campaign). If the attacker can get this additional knowledge, she/he can infer the labels by fine-tuning her/his bottom model through a further classification layer in a semi-supervised manner. This step is referred as {\em model completion} attack.
Once the training is completed, the model can predict a label for every item of the sample of the adversary.  

\subsection{Active Label Inference Attack}
\label{sub:activeAttack}
This attack is classified as {\em active} because the malicious participant performs some actions in the training stage, in particular, she/he tries to scale up the learning rate during the training phase of her/his bottom
model. In this way, she/he aims to accelerate the gradient descent on her/his bottom model to submit better features to the
server in each iteration. Consequently, she/he can force the top model to rely more on her/his bottom model than the other participants. 

Since increasing the learning rate does not always result in a more efficient gradient descent, the authors of \cite{fu2022label} perform this attack by designing and executing a malicious local optimizer. This component adaptively scales up the gradient of each parameter in the adversary's bottom model to avoid the oscillation phenomenon around
the local minimum point, that is typical of the use of an overly large learning rate for gradient descent.

Using the malicious local optimizer, the attacker can get a trained bottom model with more hidden information about labels. In addition, she/he can perform the model completion step of the passive attack (see Section \ref{sub:passiveAttack}) to fine-tune the bottom model with an additional classification layer and obtain the final label inference model. 

\subsection{Direct Label Inference Attack}
\label{sub:directAttack}

To carry out this attack the adversary directly exploits the gradients she/he receives from the top model to infer the labels of the training examples. This 
is based on the analysis of the signs of the gradients of the losses. The authors of \cite{fu2022label} demonstrate through mathematical proof that this method works for label inference in VFL without model splitting (see Section \ref{subsec:vfl} for details about model splitting).

Since no gradients are available at the inference time, with this attack, the malicious participant can only infer the labels of training examples. Nevertheless, these discovered labels can be used as the auxiliary data necessary to perform a passive label inference attack. In this way, the attacker can infer the label of an arbitrary sample.

\section{Approach Description}
\label{sec:approach}

Our approach aims to provide a countermeasure for all the types of label inference attacks described in Section \ref{sec:labelAttack}. 

\begin{figure}
    \centering
    \includegraphics[width=0.47\textwidth]{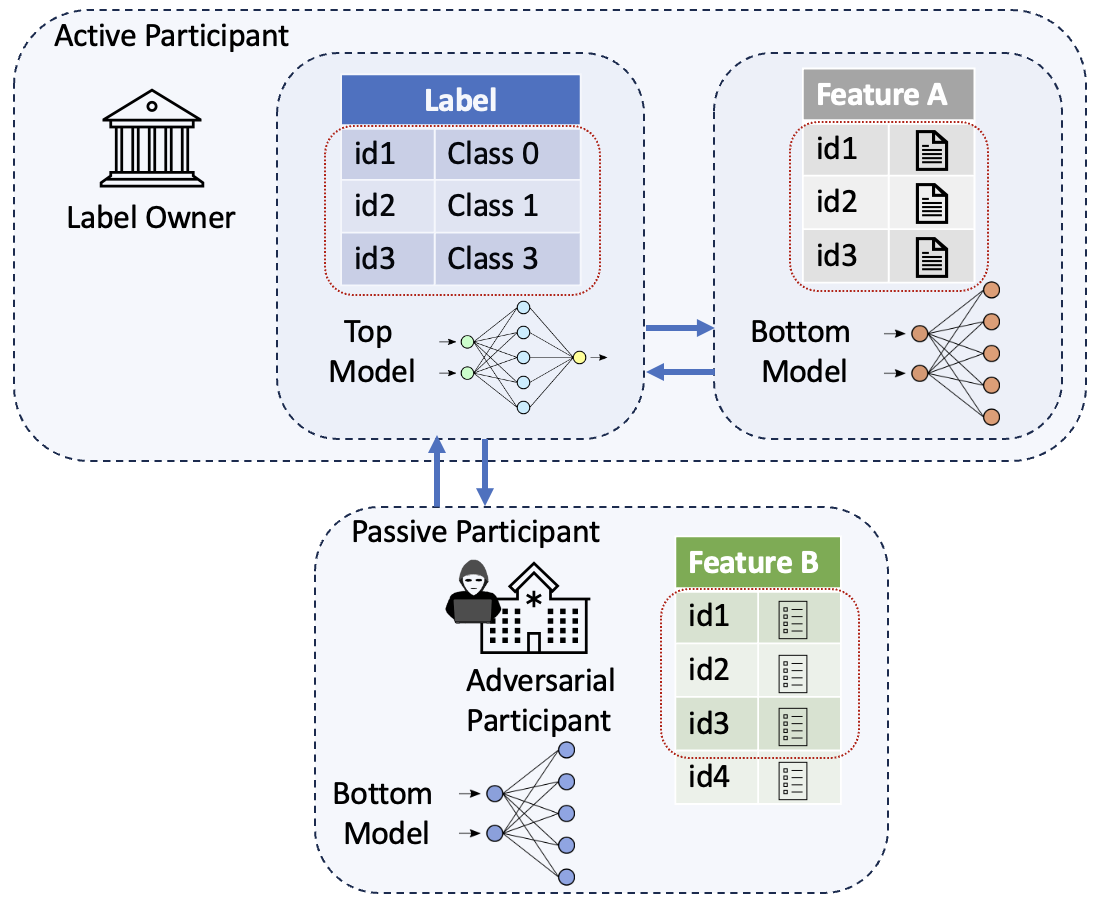}
    \caption{Label inference attack scenario against VFL\label{fig:attack}}
    \Description[]{}
\end{figure}

To better present our defense strategy, as done once again in \cite{fu2022label}, we will focus on a basic VFL attack scenario shown in Figure \ref{fig:attack}.
Here, two participants holding the same set of samples
but with features from different spaces want to train a model collaboratively through VFL.

The first participant, that is the server, runs both the top $T_M$ and the bottom $B_A$ models, hence she/he is the label owner $HL$ and holds
part of the vertically partitioned data $X_A$. For this reason, she/he is also referred to as an {\em active} client. Her/his objective is to enhance the model performance by combining her/his features with the ones of other participants coming from different business domains.

The second participant in our example is the adversarial or {\em passive} client who aims at inferring the labels from the training process and has access only to its bottom model $B_P$ and its part vertically partitioned data $X_P$. At each training round, the bottom model outputs $H=\{H_A, H_P\}$ are sent to the server running the top model $T_M$, which, hence, returns the correspondent partial gradients $\nabla H_A$ and $\nabla H_P$ of the loss $l$. These are used to update the clients' bottom model parameters $W_A$ ($W_P$). The local models updates $\nabla W_A$ and $\nabla W_P$ are calculated as follows ($CE=$ cross-entropy, $SM=$ \textit{softmax}):
\begin{gather}
    H = CONCAT(H_A, H_P), preds = T_M(H) \\
    l = CE(SM(preds),\ SM(HL)) \notag \\
    \nabla W_A =\sum \frac{\partial l}{\partial H_A} \cdot \frac{\partial H_A}{\partial B_A}. \\
    \nabla W_P =\sum \frac{\partial l}{\partial H_P} \cdot \frac{\partial H_P}{\partial B_P}. 
\end{gather}

Although the private labels $HL$ never leave the first participant's storage, the adversary can exploit the received partial gradients and the trained bottom model to conduct a label inference attack. 

In particular, to perform the first attack, or Passive Label Inference Attack, the adversary relies on a small set of auxiliary labels. If she/he manages to obtain this set she/he can fine-tune her/his bottom model through a further classification layer in a semi-supervised manner to infer the training labels.
To conduct the other attacks (i.e., the Active and the Direct Label Inference Attacks), instead, the malicious participant exploits the fact that, even though she/he does not have direct access to the label, her/his bottom model implicitly holds information about them, because of the training step. With these last strategies, the adversary cannot obtain all the private labels, but he/she can, then, run a subsequent passive attack to improve the attack performance.

At this point, we are ready to present our defense mechanism against the above-cited types of label inference attacks. In particular, we include in the active participant architecture an additional component comprised of a fine-tuned teacher network that performs a Knowledge Distillation $KD$ step to output soft labels $SL$ instead of hard ones $HL$. The output vector of a given data point contains the probabilities that it belongs to each class represented by the private labels. 
The output from this layer is then processed by an algorithm based on the concept of $k$-anonymity (see \ref{subsec:kanon} for detail) to add a second level of uncertainty. Through this further step, instead of selecting a single label for each sample, we select a set of $k$ labels in $SL$ with the highest probability. Hence, as shown in Algorithm \ref{alg:SoftLabel}, if the label is the one associated with the highest confidence it is scaled by $\epsilon$ (where  $\epsilon$ is a {\em smoothing} parameter), otherwise, if the label belongs to the $k-1$ highest probability labels (excluding the maximum) a $\epsilon/(k-1)$ factor is applied to scale up their final probability value. At this point, since the correct label for each item is obfuscated in a group of $k$ labels, as we will demonstrate in the experiments, the attacker can no longer easily infer the most probable one performing any of the above-cited attacks.
The VFL process changes as follows:
\begin{gather}
    SL \leftarrow KDk(HL, k, \epsilon) \\
    KDk(HL, k, \epsilon) = 
    \begin{cases}
        1 - \epsilon & \text{if $L_i \in \max(KD(HL))$} \\
        \frac{\epsilon}{k-1}& \text{if $L_i \in top_ k(KD(HL))$} \\
        0 & \text{otherwise}
    \end{cases} \\
    l_{kdk} = CE(SM(preds), SM(SL)) \\
    \nabla W_P =\sum \frac{\partial l_{kdk}}{\partial H_P} \cdot \frac{\partial H_P}{\partial B_P}.
\end{gather}

\noindent
Here $KD(HL)$ contains the soft labels (a probability vector) returned by the knowledge distillation model for each datapoint in the original training set.
The function $\max(KD(HL))$ returns the labels with the highest probability for each data point. Whereas, the function $top_ k(KD(HL))$ returns the set of the $k-1$ labels having the highest values following the maximum (i.e., once again, the highest probabilities of being the correct label of the target datapoint as estimated by the KD model) for each data point.

\begin{algorithm}[!ht] 
\small
\caption{Soft Label Algorithm. \label{alg:SoftLabel}}
\begin{algorithmic}[1]
\Require{\\ 
$HL$: set of Hard Labels \\ 
$K$: set of top-k labels with higher confidence \\ 
$k$: $|K|$ cardinatily of $K$ \\
$\epsilon$: smoothing parameter \\ 
$n$: number of classes}
\State $TopKIndexes$, $MaxValue$ $\gets$ $getTopKIndexes(HL, $K$)$
\State $SL$ $\gets$ $zeros(n)$
\For{$i$ in $TopKIndexes$}
    \If{$HL[i]$ == $MaxValue$}
        \State {$SL[i]$ $\gets$ $1 - \epsilon$}
    \Else
        \State {$SL[i]$ $\gets$ $\epsilon/(k-1)$}
    \EndIf
\EndFor
\end{algorithmic}
\end{algorithm}

\section{Experimental Results}
\label{sec:experimental}

In this section, we illustrate the experiments carried out to assess the performance of our defense mechanism.
Specifically, in Section \ref{sub:datasetdesc}, we describe the dataset, the evaluation metrics, and the environment used for our experiments. The remaining sections are devoted to analyzing the results and the performance of our defense approach against the different types of analyzed label inference attacks and the comparison with other defense mechanisms.

\subsection{Testbeds description}
\label{sub:datasetdesc}
To evaluate the robustness of our approach against label inference attacks we adopt some of the datasets used by \cite{fu2022label}, namely:

\begin{itemize}
    \item CIFAR-10 dataset \cite{krizhevsky2009learning} consisting of $60,000$ $32x32$ color images divided into $10$ classes with $6,000$ images per class. There are $50,000$ in training images and $10,000$ in test images.
    \item CIFAR-100 \cite{krizhevsky2009learning} dataset that is similar to CIFAR-10, but it has $100$ classes containing $600$ images each with $500$ training images and $100$ testing images per class. 
    \item CINIC-10 \cite{darlow2018cinic}, which is a large dataset and an extended alternative for CIFAR-10 with $270,000$ images, (i.e., $4.5$ times more that of CIFAR-10).
    \item Yahoo! Answers topic classification dataset \cite{zhang2015character} is formed by $10$ main categories and each class contains $140,000$ training samples and $6,000$ testing samples. 
    \item Criteo \cite{Criteo} is a real-world dataset related to commerce for predicting ad click-through rates. In this dataset, composed of only $2$ classes, both categorical and continuous features are employed.
\end{itemize}

To assess the effectiveness of our defense approach,
we adopt the following evaluation metrics:
\begin{itemize}
    \item \textbf{Top-1 Accuracy}, that is the conventional accuracy or the ratio of correctly predicted samples to the total number of samples in the dataset. It measures how many times the network has predicted the correct label with the highest probability. 
    \item \textbf{Top-5 Accuracy}, is a metric that indicates how many times the correct label appears in the network's top five predicted classes. It is useful for large-scale datasets with numerous classes and for cases in which a degree of flexibility is acceptable and the exact class can be not predicted with high confidence \cite{petersen2022differentiable}. 
    \item \textbf{Top-1 Attack Success Rate} (Top-1 ASR) that is the percentage of labels correctly extracted by attacks. 
    \item \textbf{Top-5 Attack Success Rate} (Top-5 ASR) measures how many times the label correctly extracted by attacks appears in the network's top five predicted classes. 
\end{itemize}

For our experimental campaign, we refer to an {\em Original Architecture} (OA, hereafter) that represents a VFL scenario without any defense mechanism as presented by \cite{fu2022label}. This architecture, shown in Figure \ref{fig:attack}, employs different types of networks for each of the above-described datasets. In particular, as visible in Table \ref{tab:modelArch}, the top model of the VFL is implemented through a pre-trained ResNet-18 (i.e., an 18-layer convolutional neural network pre-trained on general data and fine-tuned on the active participant data) for the CIFAR-10, CIFAR-100, and CINIC-10 datasets; a fine-tuned BERT model \cite{devlin2018bert} for Yahoo! Answer (that includes textual data); and a 3-layer Fully Connected Neural Network (FCNN-3) to process samples in the Criteo dataset. 

\begin{table}[ht]
\centering
\caption{Original Model Architectures. \label{tab:modelArch}}
\begin{tabular}{|l|c|c|}
\hline 
\textbf{Dataset} & 
\begin{tabular}[x]{@{}c@{}}\textbf{Top}\\\textbf{Model Architecture}\end{tabular} &
\begin{tabular}[x]{@{}c@{}}\textbf{Bottom}\\\textbf{Model Architecture}\end{tabular} \\
\hline
CIFAR-10 &FCNN-4 &  ResNet-18   \\
CIFAR-100 & FCNN-4 & ResNet-18  \\
CINIC-10 & FCNN-4 & ResNet-18 \\
Yahoo! Answers & FCNN-4 & Bert  \\
Criteo & FCNN-3 & FCNN-3\\
\hline
\end{tabular}
\end{table}

\begin{figure*}
    \centering
    \includegraphics[width=0.75\textwidth]{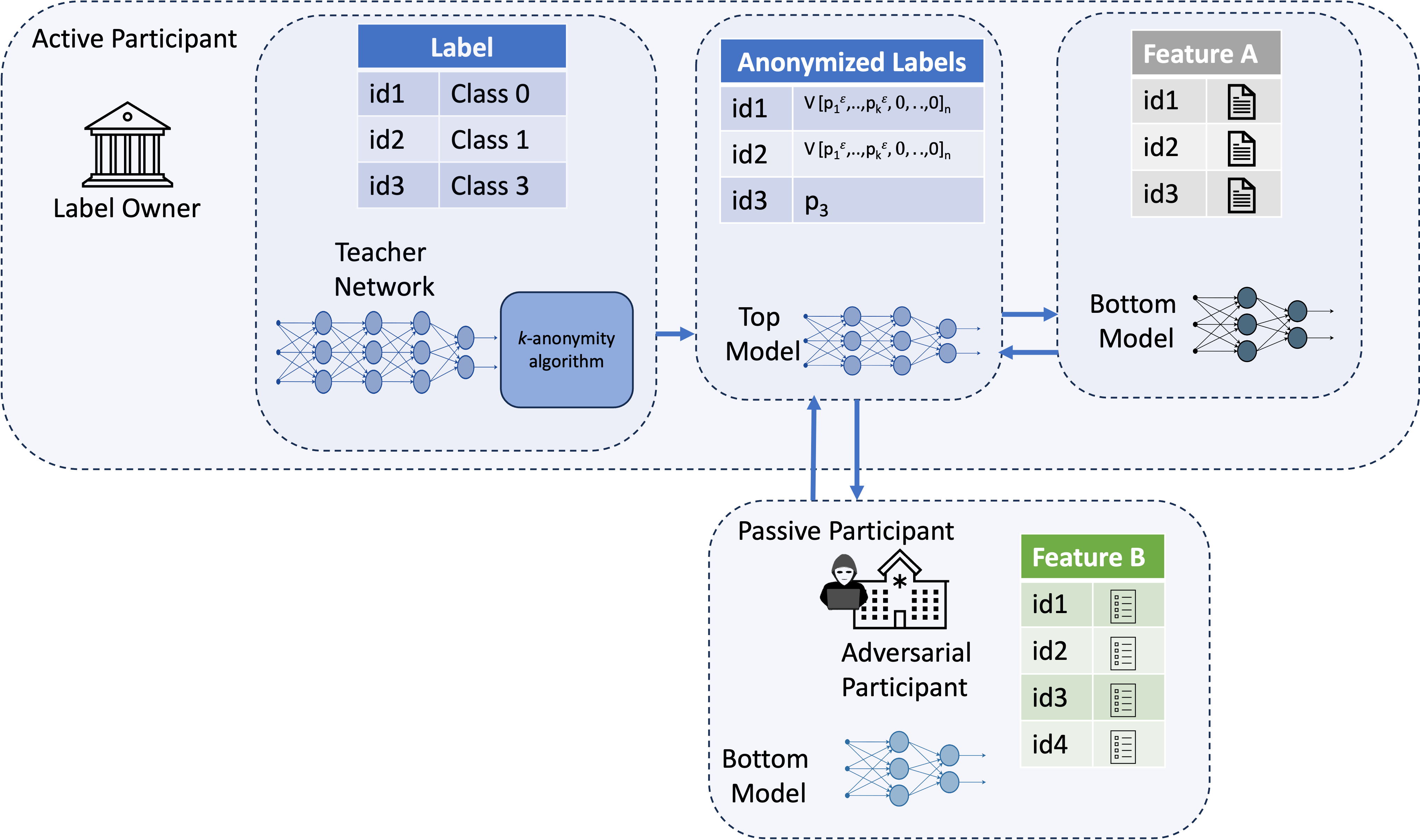}
    \caption{KD$k$ main components \label{fig:KDk}}
    \Description[]{}
\end{figure*}

Moreover, we implemented our KD$k$ solution whose components are illustrated in Figure \ref{fig:KDk}. Compared to the Original Architecture, KD$k$ includes a preliminary processing step executed only by the active participant for the anonymization of the labels. This step is realized through: {\em (i)} a teacher network that implements the Knowledge Distillation and {\em (ii)} an algorithm that obfuscates the $k$ labels with higher confidence based on $k$-anonymity. The teacher network architectures that have been implemented for each dataset is visible in Table \ref{tab:modelArchKDk}.

\begin{table}[ht]
\centering
\caption{Teacher Network Architectures for KD$k$. \label{tab:modelArchKDk}}
\begin{tabular}{|l|c|}
\hline 
\textbf{Dataset} & 
\begin{tabular}[x]{@{}c@{}}\textbf{Teacher Network}\\\textbf{Architecture}\end{tabular} \\
\hline
CIFAR-10 & ResNet-50 + FCNN-1  \\
CIFAR-100 & ResNet-50 + FCNN-1  \\
CINIC-10 & ResNet-50 + FCNN-1 \\
Yahoo! Answers & Bert + FCNN-1 \\
Criteo & FCNN-4 \\
\hline
\end{tabular}
\end{table}

All experiments have been performed on a workstation equipped
with a AMD(R) Ryzen(R) 7 CPU 5800x @ 3.80GHz, 32GB
RAM, and an NVIDIA RTX 3070Ti GPU card. 

\subsection{Label Inference Attacks Performance Comparison}
\label{sub:attackPerformance}

In this section, we report the performance results of our defense mechanism against the four types of Label Inference attacks described in Section \ref{sec:labelAttack}.

For each analyzed attack, we chose the appropriate configuration of the two anonymization parameters $\epsilon$ and $k$, where $\epsilon$ is the smoothing parameter and $k$ is the number of the highest labels considered for each data point to build our soft label solution. 

\subsubsection{Passive Label Inference Attack}
\label{sub:passiveResults}
We carried out the Passive Label Inference Attack with a $0.08\%$ of auxiliary labeled data (as proposed in \cite{fu2022label}) and we tested it against our KD$k$ framework with the two anonymization parameters $\epsilon$ and $k$ set for each dataset as are reported in Table \ref{tab:resPassive}.
This table reports also the accuracy results of the attack against the original model proposed by \cite{fu2022label} and against our defense mechanism. These values show that the performance of the attack against KD$k$ is drastically reduced and, in most cases, halved compared to the performance against the original model of \cite{fu2022label}. 
It is worth observing that, in the results obtained on the Yahoo! Answer dataset we can see a smaller reduction in the attack performance, which is caused by the fact that the bottom model is an already pre-trained Bert model. The knowledge included in the Bert model is already enough to obtain a basic classification of the text (i.e., information on the labels), even if the attacker does not infer additional information from the top model. Therefore the performance of the attack does not decrease as much as in the other cases.

\begin{table*}[ht]
\centering
\small
\caption{Passive Label Inference Attack performance against OA and KD$k$\label{tab:resPassive}}
\begin{tabular}{|l|ccc|cc|cc|}
\hline 
\multicolumn{8}{|c|}{\textbf{Attack Success Rate (ASR)}} \\
\hline 
\multirow{2}{*}{\textbf{Dataset}} & \multirow{2}{*}{\textbf{Type of ASR}} & \multirow{2}{*}{$\bm{\epsilon}$} & \multirow{2}{*}{$\bm{k}$} & \multicolumn{2}{c|}{\textbf{OA}} & \multicolumn{2}{c|}{\textbf{KD$k$}} \\
\cline{5-8}   
 &  & & & \textbf{Training Set} & \textbf{Test Set} & \textbf{Training Set} & \textbf{Test Set}\\
\hline
CIFAR-10 & Top-1 & $0.45$ & $3$ & $80.6\%$ & $61.7\%$  & $43.9\%$ & $35.8\%$ \\
CIFAR-100 & Top-1 & $0.50$ & $3$ & $31.3\%$ & $18.0\%$ & $15.9\%$ & $10.5\%$  \\
CIFAR-100 & Top-5 & $0.50$ & $3$ & $62.2\%$ & $41.0\%$ & $40.3\%$ & $29.7\%$  \\
CINIC-10 & Top-1 & $0.45$ & $3$ & $65.2\%$ & $49.0\%$  & $32.0\%$ & $26.0\%$ \\
Yahoo! Answers  & Top-1 & $0.35$ & $3$ & $63.3\%$ & $63.7\%$ & $47.5\%$ & $47.4\%$  \\
Criteo  & Top-1 & $0.40$ & $2$ & $71.2\%$ & $71.9\%$ & $50.6\%$ & $50.3\%$  \\
\hline
\end{tabular}
\end{table*}

\subsubsection{Active Label Inference Attack}
\label{sec:activeResults}
To perform the Active Label Inference Attack, we executed the malicious local optimizer in the training stage of our KD$k$ model and then we performed the completion step of the passive inference attack to get the final label inference model as done in \cite{fu2022label}.
The configurations of the two anonymization parameters $\epsilon$ and $k$ chosen for each dataset are reported in Table \ref{tab:resActive}.
Similarly to the previous experiment, when we performed an active label inference attack against our KD$k$ framework the ASR consistently decreased compared to the ASR of the attacked OA.
Observe that, the attack strategy of increasing the learning rate on the controlled client to promote more informative feedback from the server does not result in an advantage, because thanks to our defense the received signal is heavily obfuscated.
Also in this case, the results from the Yahoo! Answer dataset present a smaller reduction in the attack performance, which is, once again, caused by the fact that the bottom model is an already pre-trained Bert model. 

\begin{table*}[ht]
\centering
\small
\caption{Active Label Inference Attack performance against OA and KD$k$\label{tab:resActive}}
\begin{tabular}{|l|ccc|cc|cc|}
\hline 
\multicolumn{8}{|c|}{\textbf{Attack Success Rate (ASR)}} \\
\hline 
\multirow{2}{*}{\textbf{Dataset}} & \multirow{2}{*}{\textbf{Type of ASR}} & \multirow{2}{*}{$\bm{\epsilon}$} & \multirow{2}{*}{$\bm{k}$} & \multicolumn{2}{c|}{\textbf{OA}} & \multicolumn{2}{c|}{\textbf{KD$k$}} \\
\cline{5-8}   
 &  & & & \textbf{Training Set} & \textbf{Test Set} & \textbf{Training Set} & \textbf{Test Set}\\
\hline
CIFAR-10 & Top-1 & $0.50$ & $3$ & $84.8\%$ & $63.4\%$ & $40.5\%$ & $35.1\%$  \\
CIFAR-100 & Top-1 & $0.60$ & $3$& $39.3\%$ & $21.4\%$ & $17.6\%$ & $12.2\%$  \\
CIFAR-100 & Top-5 & $0.60$ & $3$ & $72\%$ & $47.4\%$ & $43.3\%$ & $32.7\%$  \\
CINIC-10 & Top-1 & $0.50$ & $3$ & $73.5\%$ & $50.5\%$ & $34.5\%$ & $29.3\%$  \\
Yahoo! Answers  & Top-1 & $0.40$ & $3$ & $64.2\%$ & $64.1\%$  & $52.2\%$ & $52.1\%$ \\
Criteo  & Top-1 & $0.40$ & $3$& $71.2\%$ & $71.9\%$ & $50.0\%$ & $50.0\%$  \\
\hline
\end{tabular}
\end{table*}

\subsubsection{Direct Label Inference Attack}
\label{sec:DirectResults}
We carried out the Direct Label Inference Attack described in \cite{fu2022label} and we tested it against our KD$k$ framework with the two anonymization parameters $\epsilon$ and $k$ set for each dataset as are reported in Table \ref{tab:resDirectAtt}.
In this table, we report the ASR results of the Direct Label Inference Attack against OA and our KD$k$ framework.
Since no gradients are available at the inference time, this attack can be conducted only at the training step hence we report the ASR values referred to the training set. 
As we can observe, in general, our defense mechanism can reduce the ASR of the attack by more than $60\%$ except for the Criteo dataset because the number of classes is equal to $2$ and therefore a random choice of the target label would lead to an ASR result higher than $0.5$.

\begin{table}[ht]
\centering
\begin{threeparttable}
\caption{Direct Label Inference Attack performance against OA and KD$k$ \label{tab:resDirectAtt}}
\begin{tabular}{|l|ccc|cc|}
\hline 
\multicolumn{6}{|c|}{\textbf{Attack Success Rate (ASR)}} \\
\hline 
\multirow{2}{*}{\textbf{Dataset}} & \multirow{2}{*}{\textbf{Type of ASR}}  & \multirow{2}{*}{$\bm{\epsilon}$} & \multirow{2}{*}{$\bm{k}$} & \textbf{OA} & \textbf{KD$k$}  \\
\cline{5-6}   
 &  & & & \multicolumn{2}{c|}{\textbf{Training Set}} \\
\hline
CIFAR-10 & Top-1 & $0.45$ & $3$& $100\%$ & $38.5\%$  \\
CIFAR-$100^*$ & Top-1 & $0.5$ & $3$& $100\%$ & $32.6\%$  \\
CINIC-10 & Top-1 & $0.45$ & $3$ & $100\%$ & $38.3\%$  \\
Yahoo! Answers  & Top-1 & $0.35$ & $3$& $100\%$ & $39.6\%$  \\
Criteo  & Top-1 & $0.4$ & $2$ & $100\%$ & $80\%$  \\
\hline
\end{tabular}
\begin{tablenotes}
            \item *In this case, we do not consider the Top-5 accuracy because the Top-1 is already $100\%$.
        \end{tablenotes}
\end{threeparttable}
\end{table}

\subsection{Models Performance Comparison}
\label{sub:ModelPer}

The main idea behind our approach, as presented in Section \ref{sec:approach}, is to obfuscate the information of the real label to add uncertainty in the bottom model of the attacker to inhibit the effectiveness of the attacks presented in Section \ref{sec:labelAttack}.
Inevitably, this approach will affect the performance of the model on the original task, though we try to minimize it by obfuscating the real information in a set of highly probable alternatives (and, therefore, possibly avoiding to heavily impact the performance of the top model).
The results in the previous section have been obtained by setting the anonymization parameters so to guarantee the preservation of the original global model performance.
The accuracy performance of our KD$k$ model compared to OA  (the original architecture proposed in \cite{fu2022label}), is shown in Table \ref{tab:mainTaskPerf} for each analyzed dataset.
As we can see the performance of the original model are mostly preserved with small drops of $4\%$ at maximum.
Observe that for the CIFAR-100 dataset, the accuracy result is even higher, because of the effect of KD. Indeed, for large datasets, our model benefits from the generalization capabilities of the teacher network \cite{cho2019efficacy,yang2018knowledge}.
As we said our approach can impact the model accuracy according to the strength level of the parameters $k$ and $\epsilon$. In the following Section \ref{sub:kepsilonAnalysis}, we present a detailed analysis combining parameters with different levels of strength and recording the model accuracy in model and the attack success rate.

\begin{table}[ht]
\centering
\caption{Performance on KD$k$ compared to OA for the used datasets \label{tab:mainTaskPerf}}
\begin{tabular}{|l|c|cc|}
\hline 
\multicolumn{4}{|c|}{\textbf{Model Accuracy}} \\
\hline 
\textbf{Dataset} & \textbf{Accuracy} &  \textbf{OA} & \textbf{KD$k$} \\
\hline
CIFAR-10 & Top-1 & $81\%$ & $79\%$  \\
CIFAR-100 & Top-1 & $49.1\%$ & $49.4\%$  \\
CIFAR-100 & Top-5 & $78.5\%$ & $79.9\%$  \\
CINIC-10 & Top-1 & $66.7\%$ & $64.3\%$  \\
Yahoo! Answers  & Top-1 & $71.1\%$ & $67.5\%$  \\
Criteo  & Top-1 & $71.3\%$ & $69.9\%$  \\
\hline
\end{tabular}
\end{table}

\subsection{Performance with different values of the anonymization parameters}
\label{sub:kepsilonAnalysis}
In this experiment, we analyze how both the ASR and the performance of the model (in terms of accuracy) change in relation to higher values of the anonymization parameters $\epsilon$ and $k$. For this study, we considered only three datasets, namely CIFAR-10, CIFAR-100, and CINIC-10 because they have at least $10$. Criteo has not been considered since it is a dataset with binary labels making it impossible to test our solution with $k$ higher than 2. Yahoo instead is not included since relies on a pre-trained Bert model and, as we already stated in Section \ref{sub:passiveResults}, its accuracy is intrinsecally guarateed by the performance of such underling model, hence it cannot decrease lower than the values presented in Table \ref{tab:resPassive} with any parameter combination.

That said, we studied the performance for different values of $k$ (i.e., $k=3$, $k=5$, and $k=10$). As typically done in the literature \cite{fu2022label, liu2022batch}, for CIFAR-100 we consider only the Top-5 accuracy that provides a more nuanced evaluation because of the large number of classes.

\begin{figure*}[ht]
    \centering
    \includegraphics[width=0.63\textwidth]{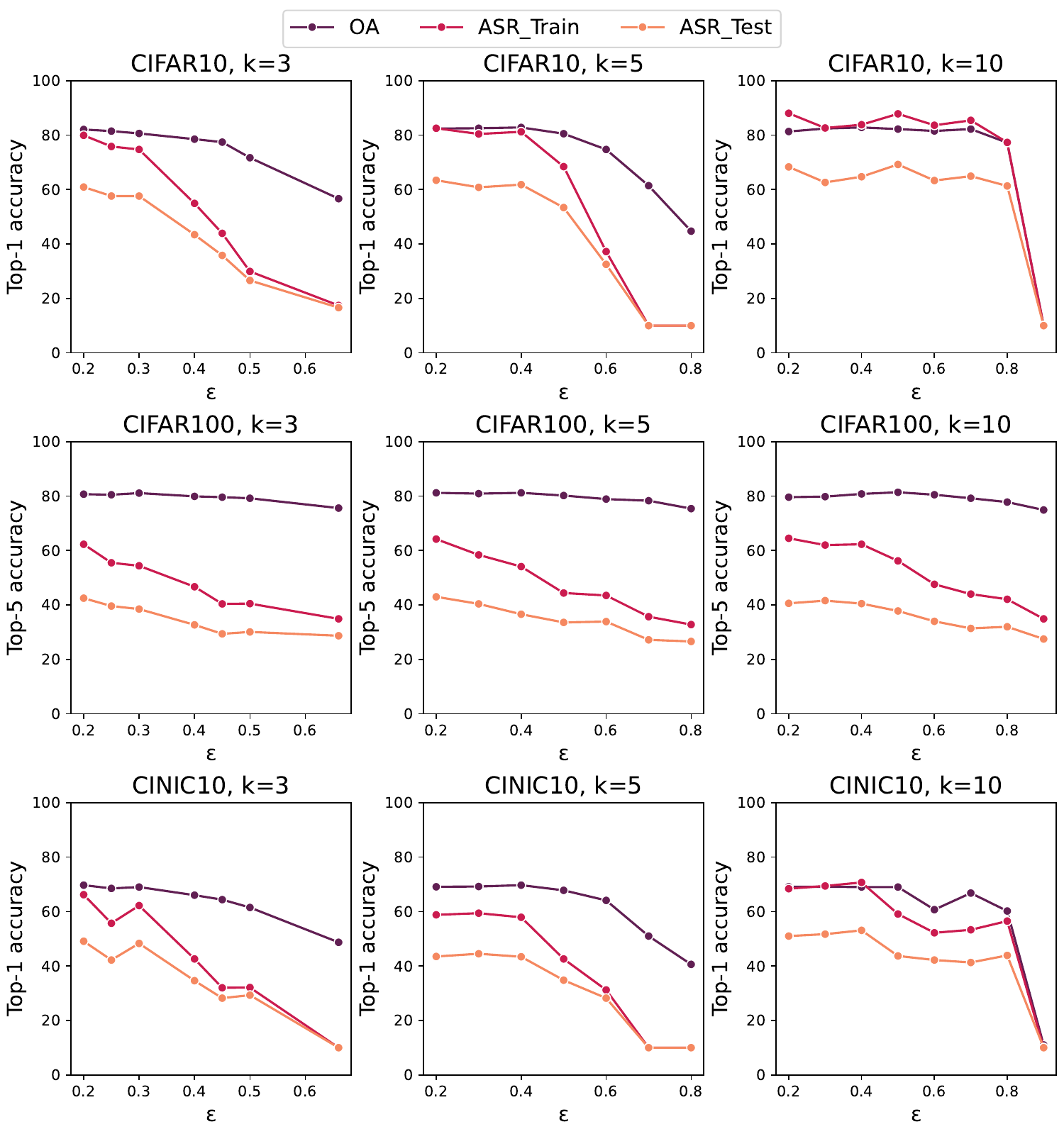}
    \caption{Analysis of the performance of the attack and the performance of KD$k$ for different $\epsilon$ and $k$ values \label{fig:changingAnonParams}}
    \Description[]{}
\end{figure*}

As visible in Figure \ref{fig:changingAnonParams}, the use of higher $\epsilon$ values results in balancing the probabilities of the classes, and this affects the overall performance of both the attack and the KD$k$ model.
Using different $k$ values, instead, does not affect our defense mechanism.
Interestingly, setting $k=10$ and using a dataset composed of $10$ classes make the defense ineffective.
This result confirms our intuition behind the logic that makes our proposal work.
The reason why our approach is effective relies on the uncertainty instilled in the bottom models obtained by anonymizing the real label between $k$ additional and related ones (as indicated by our knowledge distillation component). In the case of $k=10$, we are setting all the secondary labels to the same probability. This breaks the main logic behind our approach.
Setting all the secondary labels to the same value produces similar (with the addition of an offset) loss values compared to the scenario using hard labels, directly. In this case, the offset added to the cross-entropy loss is not sufficient to properly obfuscate the labels, thus resulting in a small decrease in the accuracy of the attack in the case of the CINI10 dataset or can even be ineffective in the case of CIFAR-10.
To be effective with $k=10$, our approach must push the $\epsilon$ value to extreme values.
This setting is effective against the attack but also prevents the model from training using the probability distribution balanced across all the labels, thus resulting in an accuracy close to random guessing. 

\subsection{Comparison with other Defense Mechanisms}

\begin{figure*}
    \centering
    \includegraphics[width=0.97\textwidth]{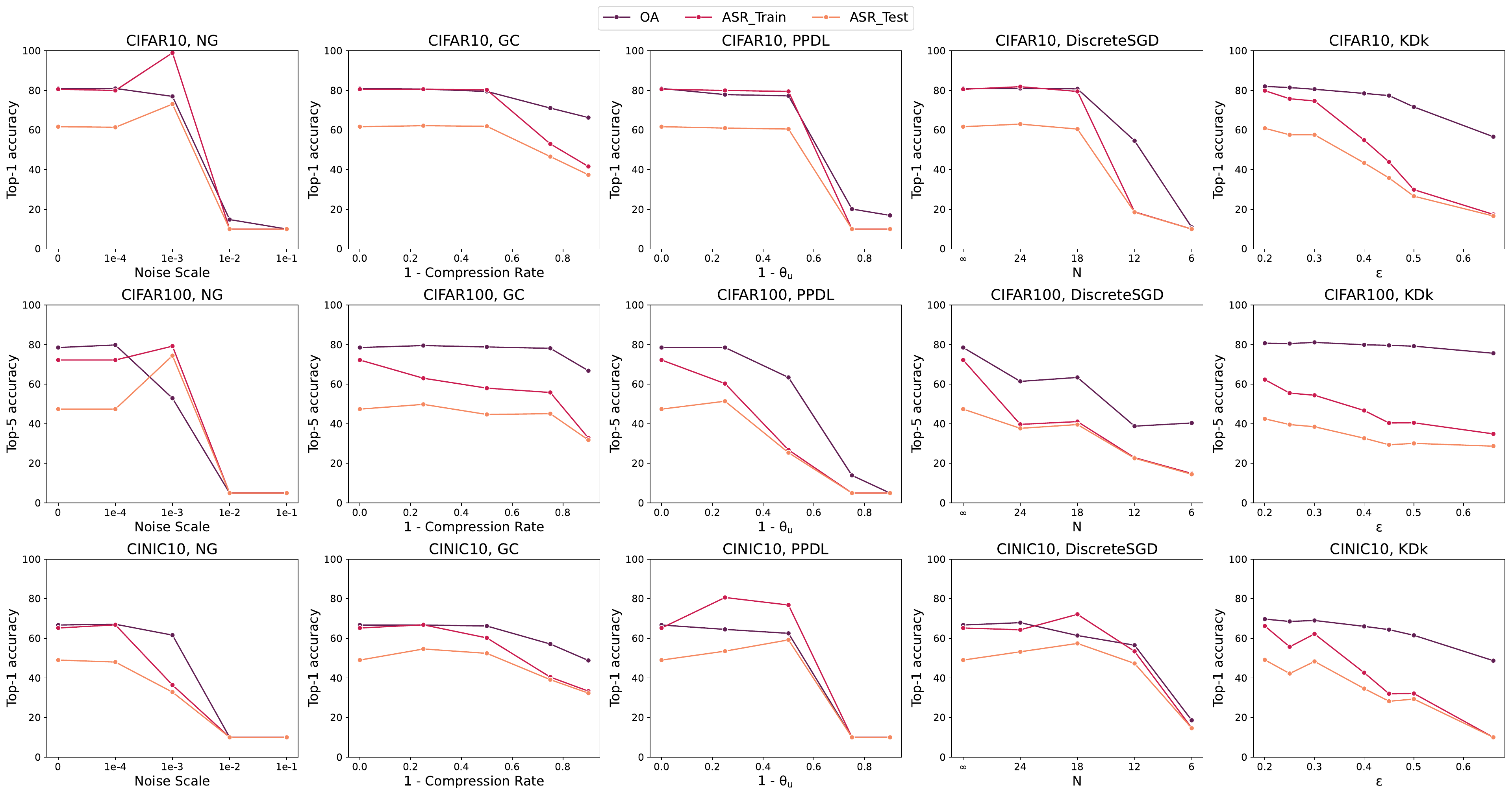}
    \caption{Comparison with other Defense Mechanisms against passive and active label inference attacks \label{fig:comparisonDefenseMC}}
    \Description[]{}
\end{figure*}

\begin{table*}[ht]
\centering
\scriptsize
\caption{Comparison with other Defense Mechanisms against direct label inference attacks using CIFAR datasets \label{tab:comparisonDefenseDACifar}}
\begin{tabular}{|c|cc|cc|cc|}
\hline
\multicolumn{3}{|c}{ } & \multicolumn{2}{|c|}{\textbf{CIFAR-10}} & \multicolumn{2}{c|}{\textbf{CIFAR-100}}\\
\hline
 \multirow{5}{*}{\begin{tabular}[x]{@{}c@{}}\textbf{Noisy}\\\textbf{Gradients}\end{tabular} } &\multicolumn{2}{c|}{\textbf{Noise Scale}} & \textbf{Model Accuracy} & \textbf{Attack Accuracy} & \textbf{Model Accuracy} & \textbf{Attack Accuracy} \\
\cline{2-7}
& \multicolumn{2}{c|}{1e-4} & $81.4\%$ & $80.6\%$ & $82.4\%$ & $12.2\%$ \\
& \multicolumn{2}{c|}{1e-3} & $81.1\%$ & $49.1\%$ & $83.1\%$ & $2.0\%$ \\
& \multicolumn{2}{c|}{1e-2} & $71.9\%$ & $24.5\%$ & $5.1\%$ & $2.0\%$ \\
& \multicolumn{2}{c|}{1e-1} & $10.0\%$ & $12.7\%$ & $5.0\%$ & $0.6\%$ \\
\hline
 \multirow{5}{*}{\begin{tabular}[x]{@{}c@{}}\textbf{Gradient}\\\textbf{Compression}\end{tabular} } &\multicolumn{2}{c|}{\textbf{Compression Rate}} & \textbf{Model Accuracy} & \textbf{Attack Accuracy} & \textbf{Model Accuracy} & \textbf{Attack Accuracy} \\
\cline{2-7}
& \multicolumn{2}{c|}{$75\%$}  & $80.4\%$ & $99.9\%$ & $82.4\%$ & $100\%$\\
& \multicolumn{2}{c|}{$50\%$}  & $80.5\%$ & $99.3\%$ & $83.1\%$ & $100\%$\\
& \multicolumn{2}{c|}{$25\%$}  & $78.4\%$ & $92.4\%$ & $82.4\%$ & $99.9\%$\\
& \multicolumn{2}{c|}{$10\%$}  & $10.0\%$ & $0.1\%$  & $73.8\%$ & $99.9\%$ \\
\hline
 \multirow{5}{*}{\begin{tabular}[x]{@{}c@{}}\textbf{Privacy-}\\\textbf{preserving
DL}\end{tabular}} &\multicolumn{2}{c|}{\textbf{$\Theta_u$}} & \textbf{Model Accuracy} & \textbf{Attack Accuracy} & \textbf{Model Accuracy} & \textbf{Attack Accuracy}  \\
\cline{2-7}
& \multicolumn{2}{c|}{$0.75$}  & $79.8\%$ & $39.0\%$ & $81.9\%$ & $4.6\%$  \\
& \multicolumn{2}{c|}{$0.50$}  & $80.5\%$ & $38.9\%$ & $81.7\%$ & $4.5\%$  \\
& \multicolumn{2}{c|}{$0.25$}  & $19.9\%$ & $0.1\%$  & $5.2\%$ & $1.1\%$  \\
& \multicolumn{2}{c|}{$0.10$}  & $10.0\%$ & $0.1\%$  & $5.0\%$ & $0.9\%$ \\
\hline
 \multirow{5}{*}{\begin{tabular}[x]{@{}c@{}}\textbf{Discrete}\\\textbf{SGD}\end{tabular} } &\multicolumn{2}{c|}{\textbf{N}} & \textbf{Model Accuracy} & \textbf{Attack Accuracy} & \textbf{Model Accuracy} & \textbf{Attack Accuracy} \\
\cline{2-7}
& \multicolumn{2}{c|}{24} & $81.0\%$ & $96.7\%$ & $11.2\%$ & $99.9\%$\\
& \multicolumn{2}{c|}{18} & $80.8\%$ & $94.3\%$ & $8.8\%$ & $99.9\%$ \\
& \multicolumn{2}{c|}{12} & $78.7\%$ & $94.7\%$ & $7.1\%$ & $99.9\%$  \\
& \multicolumn{2}{c|}{6} & $74.3\%$  & $91.5\%$ & $7.3\%$ & $99.7\%$ \\
\hline
 \multirow{5}{*}{\textbf{KD$k$}} &\textbf{k} &$\bm{\epsilon}$ & \textbf{Model Accuracy} & \textbf{Attack Accuracy} & \textbf{Model Accuracy} & \textbf{Attack Accuracy} \\
\cline{2-7}
& $3$  & $0.45$ & $79.0\%$ & $38.5\%$ & $80.6\%$ & $32.6\%$\\
& $5$  & $0.70$ & $71.1\%$ & $23.0\%$ & $80.5\%$ & $19.2\%$\\
& $5$  & $0.75$ & $66.7\%$ & $21.7\%$ & $79.7\%$ & $19.1\%$  \\
& $5$  & $0.85$ & $37.5\%$ & $14.2\%$ & $76.7\%$ & $18.8\%$ \\
\hline
\end{tabular}%

\end{table*}

\begin{table}[ht]
\centering
\scriptsize
\caption{Comparison with other Defense Mechanisms against direct label inference attacks using CINIC-10 dataset\label{tab:comparisonDefenseDACinic}}
\begin{tabular}{|c|cc|cc|}
\hline
\multicolumn{3}{|c}{ } & \multicolumn{2}{|c|}{\textbf{CINIC-10}}\\
\hline
 \multirow{5}{*}{\begin{tabular}[x]{@{}c@{}}\textbf{Noisy}\\\textbf{Gradients}\end{tabular} } &\multicolumn{2}{c|}{\textbf{Noise Scale}} & \textbf{Model Accuracy} & \textbf{Attack Accuracy} \\
\cline{2-5}
& \multicolumn{2}{c|}{1e-4} & $70.5\%$ & $84.3\%$\\
& \multicolumn{2}{c|}{1e-3} & $69.9\%$ & $49.7\%$ \\
& \multicolumn{2}{c|}{1e-2} & $55.5\%$ & $24.3\%$\\
& \multicolumn{2}{c|}{1e-1} & $10.3\%$ & $12.6\%$\\
\hline
 \multirow{5}{*}{\begin{tabular}[x]{@{}c@{}}\textbf{Gradient}\\\textbf{Compression}\end{tabular} } &\multicolumn{2}{c|}{\textbf{Compression Rate}} & \textbf{Model Accuracy} & \textbf{Attack Accuracy} \\
\cline{2-5}
& \multicolumn{2}{c|}{$75\%$}  &$70.9\%$ & $99.8\%$ \\
& \multicolumn{2}{c|}{$50\%$}  & $69.1\%$ & $99.3\%$ \\
& \multicolumn{2}{c|}{$25\%$}  & $54.7\%$ & $92.5\%$ \\
& \multicolumn{2}{c|}{$10\%$}  & $10.0\%$ & $0.01\%$ \\
\hline
 \multirow{5}{*}{\begin{tabular}[x]{@{}c@{}}\textbf{Privacy-}\\\textbf{preserving
DL}\end{tabular}} &\multicolumn{2}{c|}{\textbf{$\Theta_u$}} & \textbf{Model Accuracy} & \textbf{Attack Accuracy} \\
\cline{2-5}
& \multicolumn{2}{c|}{$0.75$}  & $69.4\%$ & $38.9\%$ \\
& \multicolumn{2}{c|}{$0.50$}  & $68.4\%$ & $38.6\%$ \\
& \multicolumn{2}{c|}{$0.25$}  & $20.8\%$ & $0.10\%$ \\
& \multicolumn{2}{c|}{$0.10$}  & $12.9\%$ & $0.04\%$ \\
\hline
 \multirow{5}{*}{\begin{tabular}[x]{@{}c@{}}\textbf{Discrete}\\\textbf{SGD}\end{tabular} } &\multicolumn{2}{c|}{\textbf{N}} & \textbf{Model Accuracy} & \textbf{Attack Accuracy} \\
\cline{2-5}
& \multicolumn{2}{c|}{24} & $63.1\%$ & $97.9\%$ \\
& \multicolumn{2}{c|}{18} & $59.6\%$ & $95.6\%$ \\
& \multicolumn{2}{c|}{12} & $45.8\%$ & $94.3\%$ \\
& \multicolumn{2}{c|}{6}  & $43.6\%$ & $90.3\%$ \\
\hline
 \multirow{5}{*}{\textbf{KD$k$}} &\textbf{k} &$\bm{\epsilon}$ & \textbf{Model Accuracy} & \textbf{Attack Accuracy} \\
\cline{2-5}
& $3$  & $0.45$ & $67.7\%$ & $38.3\%$ \\
& $5$  & $0.70$ & $62.2\%$ & $24.1\%$ \\
& $5$  & $0.75$ & $56.7\%$ & $22.2\%$ \\
& $5$  & $0.85$ & $34.5\%$ & $15.3\%$ \\
\hline
\end{tabular}%

\end{table}

In the proposal of \cite{fu2022label} several defensive strategies are applied to the gradients to
prevent information leakage from the server and try to mitigate the different label inference attacks.
In this section we compare our defense mechanisms with the following approaches in \cite{fu2022label}:

\begin{itemize}
    \item \textbf{Noisy Gradients (NG)}. To perform this defense in  VFL, the server adds a laplacian noise to gradients before sending them to passive participants. The metric we analyze to compare this approach with our KD$k$ is the {\em noise scale}, which represents several scales of the used laplacian noise.
    \item \textbf{Gradient Compression (GC)}. This strategy used for communication efficiency and privacy protection consists of sharing fewer gradients with the largest absolute values. The metric we consider to compare this approach with our KD$k$ is the {\em compression rate}, which is the ratio between the uncompressed size and compressed size of the gradient values.
    \item \textbf{Privacy-Preserving Deep Learning (PPDL)}. In each iteration, the server {\em(i)} randomly selects one gradient value and adds noise to this gradient; {\em(ii)} sets to zero the gradient values smaller than a threshold value $\tau$;  {\em (iii)} repeats the first two steps until $\Theta_u$ fraction of gradient values are collected. Both $\tau$ and $\Theta_u$ are hyperparameters to balance the trade-off between model performance and defense performance. We evaluate the performance of this type of defense by analyzing different settings of the hyperparameter $\Theta_u$.
    \item \textbf{DiscreteSGD} a customized version of signSGD \cite{bernstein2018signsgd} thought for VFL. The defense mechanism proceeds as follows. {\em (i)} In the first epoch, the server observes the distribution of the shared gradients. Following the three-sigma rule \cite{pukelsheim1994three}, the server sets an interval as $[\mu - 2\sigma, \mu+2\sigma]$ (where $\mu$ is the mean and $\sigma$ is the standard deviation). The gradients outside of the interval are regarded as outliers and not considered. {\em (ii)} The server slices the interval into N sub-intervals. {\em (iii)} Before transmitting the gradients to all the participants, the server first rounds each gradient value to the nearest endpoint of the sub-intervals. The hyperparameter $N$ controls how much magnitude information of the shared gradients is preserved. 
\end{itemize}

We evaluate the four defense approaches introduced above and we compare them with our KD$k$ approach. 

For this experiment, we performed both passive and active label inference attacks on three datasets: CIFAR-10, CIFAR-100, and CINIC-10. Observe that, once again, as typically done in the related literature, for CIFAR-100 we considered only Top-5 accuracy to cope with the large number of classes.
As for the setting of the different defense mechanisms, we considered the following parameters: Laplacian noise level $\in \{10^{-1}, 10^{-2}, 10^{-3},10^{-4}\}$, gradient compression percentage $\in \{75\%, 50\%, 25\%, 10\%\}$, PPDL $\Theta_u$ fraction $\in \{10\%, 25\%, 50\%, 75\% \}$, DisctreteSGD number of intervals N $\in \{6, 12, 18, 24\}$. The parameters of our approach instead are set as follows: $k=3$ and $\epsilon$ varying between the values $\in \{0.25, 0.3, 0.45, 0.5, 0.66\}$.

\subsubsection{Passive and Active Attacks}
The results of the defenses against the model completion attack are reported in Figure \ref{fig:comparisonDefenseMC}.
As visible in Figures \ref{fig:comparisonDefenseMC}, for noisy gradients (NG), we experimented using several scales of laplacian noise to evaluate its
defense performance against model completion inference attack. Obviously the greater the value of the noise scale the more successful are all the mitigation techniques. To be effective this defense must apply to the gradients an extremely high level of noise that disrupts the performance of the model on the original task. With lower levels of noise, it is interesting to see how this approach can even help obtaining higher attack performance. 

In the second column of sub-figures in Figure \ref{fig:comparisonDefenseMC}, we evaluate Gradient Compression (GC) techniques for different compression rates. Also from these figures, we can notice that for greater compression rates both the model and the attack performance decrease. 
We can see how between the selected defenses compared to ours, gradient compression is the best preserving the original accuracy of the model but only slightly affecting the performance of the attack, especially with lower values of compression.
As for the PPDL mechanism, from the sub-figures in the third column in Figure \ref{fig:comparisonDefenseMC} we can notice that for all three analyzed datasets, the defense can mitigate label inference attacks with the hyperparameter $\Theta_u$ set to $0.25$ or lower (i.e., the accuracy result is lower than $40\%$ for $1-\Theta_u = 0.75$).
Even in this case, we can notice how the defense is effective only with a high level of manipulation of the gradient resulting in a heavy loss in terms of performance on the original task for both CIFAR-10 and CINIC-10. As for CIFAR-100, instead, PPDL represents the best-performing defense we are comparing with.

Finally, similarly to the previous defenses,  we can notice how DiscreteSGD is not capable of affecting the attack preserving the functionality of the original model. This defense can achieve slightly high per formance only for CIFAR-10.

Looking at our solution compared to the others we can see how we can prevent the attack decreasing its success rate to almost the same as a random guess value on CIFAR-10 and CINIC-10 using extreme values for $\epsilon$, while preserving most of the accuracy of the main model.
It is also interesting to see how, even with lower defense intensity values, our approach affects the attack still more than the other solutions. 

\subsubsection{Direct Label Inference Attack}
In Tables \ref{tab:comparisonDefenseDACifar} and \ref{tab:comparisonDefenseDACinic}, we analyze the performance of the three analyzed defense mechanisms and KD$k$ against the direct label inference attack. 
The employed datasets are, once again, CIFAR-10, CIFAR-100 (see Table \ref{tab:comparisonDefenseDACifar}), and CINIC-10 (see Table \ref{tab:comparisonDefenseDACinic}).
As we can see, the behavior of the defenses we are comparing with is similar to the one witnessed for the passive and active attacks. Indeed, especially for CIFAR-10 and CINIC-10, the defenses are effective only when the alteration is such that the impact on the main task accuracy is not negligible.
The only countermeasure capable of matching our solution in terms of preservation of the original model accuracy and detriment of the attack performance is the Noisy Gradients defense. Looking at CIFAR-100, instead, we can see how also the PPDL defense is capable of achieving good results. 
Our defense compared to the others is equally effective on the three considered datasets. In this case, though, a more powerful setting is required to counter the more powerful Direct Attack.

In summary, from the above experiments we can conclude that, our defense strategy is the only one obtaining good performance against all the different attacks and for all the analyzed datasets.
Generally, most of the other defeses failed protecting against label inference attacks. Only the PPDL and Noisy Gradients succeeded in {\em some} of the considered attack scenarios but, as visible in our results, they cannot be used as a general defense because they do not provide an adequate protection against all the possible attack settings.

\section{Conclusion}
\label{sec:conclusion}

Federated Learning (FL) is a novel paradigm aiming at training ML models in a privacy-preserving and collaborative way. Differently from Horizontal FL, in Vertical FL (VFL) participants share the same sample space, but their local private data differ in the feature space. Moreover, in standard VFL, the labels of the samples are sensitive information and should be protected from honest-but-curious parties. Hence, only the aggregating server (or active
actor) knows them, whereas they are kept secret from all the other parties (passive actors). Nevertheless, recent works have started to describe label leakage issues in this context proposing strategies for label inference attacks, namely passive, active, and direct attacks.
In this paper, we analyzed such existing attacks and proposed a novel defense mechanism, called KD$k$, able to protect VFL from all the known types of label inference attacks with very high performance. Our approach modifies the active participant model, integrating both a Knowledge Distillation teacher network and a $k$-anonymity processing step to obtain a group of $k$ most probable soft labels for each item instead of a single hard label. This adds a level of uncertainty that prevents the attacker from performing label inference successfully. 
We tested the performance of our solution with a thorough experimental campaing, whose objective was to demostrate that our approach can effectively inhibit the attacker from being able to perform label inference (attack success rate reduced, in some cases, even more than 60\% with respect to its performance in the absence of our defense), still maitaining an almost unaltered accuracy of the federated global model (less than 2\% performance decrease on average).
Finally, we demonstrated the superiority of our proposal with respect to the most recent and state-of-the-art existing defenses, which proved to be either uneffective agaist the attack, or, in some cases, effective against only some attack variants and often at the cost of an extremely high, and hence not acceptable, impact on the federated global model performance.

The proposal and results described in this paper must not be seen as the final conclusion of this reasearch. As a matter of fact, in the future, we plan to further develop our proposed KD$k$ defense method to provide enhanced protection for other kinds of FL and attacks, designing a more complete protection framework. For instance, we intend to focus also on Horizontal FL. Due to the peculiarities of this variant, a thorough examination must be conducted to comprehend how our defense mechanism can be adjusted according to it.




\end{document}